\documentclass{article}

\usepackage[usenames,dvipsnames,svgnames,table]{xcolor}
\usepackage[breaklinks=true,hyperfootnotes=false,colorlinks,bookmarks=false,citecolor=MidnightBlue,linkcolor=BrickRed,urlcolor=MidnightBlue]{hyperref}
\usepackage{nips13submit_e,times}
\usepackage[round]{natbib}
\usepackage{amsmath}
\usepackage{graphicx}
\usepackage{tikz}
\usepackage{multirow}
\usepackage{colortbl}

\graphicspath{{figures/}}

\usetikzlibrary{calc}

\DeclareUrlCommand\emailurl{}
\newcommand{\email}[1]{\href{mailto:#1}{\emailurl{#1}}}

\definecolor{Red}{rgb}{1,0,0}
\definecolor{Green}{rgb}{0,0.69,0}
\definecolor{Blue}{rgb}{0,0,1}
\definecolor{LightBlue}{rgb}{0,0.5,1}
\definecolor{lightBlue}{rgb}{0,0.5,1}
\definecolor{Skin}{rgb}{1,0.71,0.69}
\definecolor{Grey}{rgb}{0.5,0.5,0.5}
\definecolor{LightGrey}{rgb}{0.6,0.6,0.6}
\definecolor{Black}{rgb}{0,0,0}
\definecolor{White}{rgb}{1,1,1}
\newcommand{\red}{\color{Red}}

\newcommand{\lightBlue}{\color{LightBlue}}

\newcommand{\eg}{\emph{e.g.},}

\newcommand{\vs}{\emph{vs.}}
\newcommand{\define}{\stackrel{\triangle}{=}}

\title{Saying What You're Looking For:\\Linguistics Meets Video Search}

\author{%
  Andrei Barbu\thanks{School of Electrical and Computer Engineering, Purdue
    University, West Lafayette IN 47907-2035}\\
  \email{andrei@0xab.com}\\\And
  N. Siddharth\footnotemark[1]
  \\\email{siddharth@iffsid.edu}\\\And
  Jeffrey Mark Siskind\footnotemark[1]
  \\\email{qobi@purdue.edu}
}

\nipsfinalcopy 

\begin{document}

\maketitle

\begin{abstract}
  We present an approach to searching large video corpora for video clips
  which depict a natural-language query in the form of a sentence.
  This approach uses compositional semantics to encode subtle meaning
  that is lost in other systems, such as the difference between two
  sentences which have identical words but entirely different meaning:
  \emph{The person rode the horse} \vs\ \emph{The horse rode the
    person}.
  Given a video-sentence pair and a natural-language parser, along with
  a grammar that describes the space of sentential queries, we produce
  a score which indicates how well the video depicts the sentence.
  We produce such a score for each video clip in a corpus and return a
  ranked list of clips.
  Furthermore, this approach addresses two fundamental problems
  simultaneously: detecting and tracking objects, and recognizing
  whether those tracks depict the query.
  Because both tracking and object detection are unreliable, this uses
  knowledge about the intended sentential query to focus the tracker
  on the relevant participants and ensures that the resulting tracks
  are described by the sentential query.
  While earlier work was limited to single-word queries which
  correspond to either verbs or nouns, we show how one can search for
  complex queries which contain multiple phrases, such as
  prepositional phrases, and modifiers, such as adverbs.
  We demonstrate this approach by searching for 141 queries involving
  people and horses interacting with each other in 10 full-length
  Hollywood movies.
\end{abstract}

\section{Introduction}

Video search engines lag behind text search engines in their wide use
and performance.
This is in part because the most attractive interface for finding
videos remains a natural-language query in the form of a sentence but
determining if a sentence describes a video remains a difficult task.
This task is difficult for a number of different reasons: unreliable
object detectors which are required to determine if nouns occur,
unreliable event recognizers which are required to determine if verbs
occur, the need to recognize other parts of speech such as adverbs or
adjectives, and the need for a representation of the semantics of a
sentence which can faithfully encode the desired natural-language
query.
We propose an approach which simultaneously addresses all of these
problems.
Systems to date generally attempt to independently address the various
aspects that make this task difficult.
For example, they attempt to separately find videos that depict nouns and videos
that depict verbs and essentially take the intersection of the two sets of
videos.
This general approach of solving these problems piecemeal cannot
represent crucial distinctions between otherwise similar input
queries.
For example, if you search for \emph{The person rode the horse} and
for \emph{The horse rode the person}, existing systems would give the
same result for both queries as they each contain the same words, but
clearly the desired output for these two queries is very different.
We develop a holistic approach which both combines tracking and word
recognition to address the problems of unreliable object detectors and
trackers and at the same time uses compositional semantics to
construct the meaning of a sentence from the meaning of its words in order
to make crucial but otherwise subtle distinctions between otherwise
similar sentences.
Given a grammar and an input sentence, we parse that sentence and, for
each video clip in a corpus, we simultaneously track all objects that the
sentence refers to and enforce that all tracks must be described by
the target sentence using an approach called the \emph{sentence
  tracker}.
Each video is scored by the quality of its tracks, which are
guaranteed by construction to depict our target sentence, and the
final score correlates with our confidence that the resulting tracks
correspond to real objects in the video.
We produce a score for every video-sentence pair and return multiple
video hits ordered by their scores.

\citet{hu2011survey} note that recent work on semantic video
search focuses on detecting nouns, detecting verbs, or using language
to search already-existing video annotation.
Work that detects objects does not employ object detectors, but
instead relies on statistical features to cluster videos with similar
objects.
\citet{sivic2003video} extract local features from a positive example
of an object to find key frames that contain the same object.
\citet{anjulan2009unified} track stable image patches to extract
object tracks over the duration of a video and group similar tracks
into object classes.
Without employing an object detector, these methods cannot search a
collection of videos for a particular object class but instead must
search by example.

Prior work on verb detection does not integrate with work on object
detection.
%
%
\citet{chang2002extract} find one of four different highlights in
basketball games using hidden Markov models and the expected structure
of a basketball game.
It does not detect objects but instead classifies entire presegmented
clips, is restricted to a small number of domain-specific actions, and
supports only single-word queries.
\citet{yu2003trajectory} track one object, a soccer ball, and detect
actions being performed on that object during a match by the
position and velocity of the object.
It supports a small number of domain-specific actions and is limited
to a single object.
In summary, the above approaches only allow for searching for a single
word, a verb, and are domain-specific.

Prior work on more complex queries involving both nouns and verbs
essentially encodes the meaning of a sentence as a conjunction of
words, discarding the semantics of the sentence.
\citet{christel2004exploiting}, \citet{worring2007mediamill}, and
\citet{snoek2007learned} present various combinations of text search,
verb retrieval, and noun retrieval, and essentially allow for finding
videos which are at the intersection of multiple search mechanisms.
\citet{aytar2008utilizing} rely on annotating a video corpus with
sentences that describe each video in that corpus.
They employ text-based search methods which given a query, a
conjunction of words, attempt to find videos of similar concepts as
defined by the combination of an ontology and statistical features
of the videos.
Their model for a sentence is a conjunction of words where
higher-scoring videos more faithfully depict each individual word but
the relationship between words is lost.
None of these methods attempt to faithfully encode the semantics of a
sentence and none of them can encode the distinction between \emph{The
  person hit the ball} and \emph{The ball hit the person}.

In what follows, we describe a system, which unlike previous approaches,
allows for a natural-language query of video corpora which have no
human-provided annotation.
Given a sentence and a video corpus, it retrieves a ranked list of
videos which are described by that sentence.
We show a method of constructing a lexicon with a small number of
parameters, which are reused among multiple words, making training those
parameters easy and ensuring the system need not be shown positive
examples of every word in the lexicon.
We present a novel way to combine the semantics of words into the
semantics of sentences and to combine sentence recognition with object
tracking in order to score a video-sentence pair.
To demonstrate this approach, we run 141 natural-language queries of a
corpus of 10 full-length Hollywood movies using a grammar which
includes nouns, verbs, adverbs, and spatial-relation and motion prepositions.
This is the first approach which can search for complex queries which
include multiple phrases, such as prepositional phrases, and modifiers,
such as adverbs.

\section{Tracking}

To search for videos which depict a sentence, we must first track
objects that participate in the event described by that sentence.
Tracks consist of a single detection per frame per object.
To recover these tracks, we employ detection-based tracking.
An object detector is run on every frame of a video producing a set of
axis-aligned rectangles along with scores which correspond to the
strength of each detection.
There are two reasons why we need a tracker and cannot just take the
top-scoring detection in every frame.
First, there may be multiple instances of the same object in the field
of view.
Second, object detectors are extremely unreliable.
Even on standard benchmarks, such as the PASCAL Visual Object Classes
(VOC) Challenge, even the best detectors for the easiest-to-detect
object classes achieve average-precision scores of 40\% to 50\%
\citep{Everingham10}.
We overcome both of these problems by integrating the intra-frame
information available from the object detector with inter-frame
information computed from optical flow.

We expect that the motion of correct tracks agrees with the motion of
the objects in the video which we can compute separately and
independently of any detections using optical flow.
We call this quantity the motion coherence of a track.
In other words, given a detection corresponding to an object in the
video, we compute the average optical flow inside that detection and
forward-project the detection along that vector, and expect to find
a strong detection in the next frame at that location.
We formalize this intuition into an algorithm which finds an optimal
track given a set of detections in each frame.
Each detection $j$ has an associated axis-aligned rectangle~$b^t_j$
and score~$f(b^t_j)$ and each pair of detections has an associated
temporal coherence score~$g(b^{t-1}_{j^{t-1}},b^t_{j^t})$ where $t$ is
the index of the current frame in a video of length $T$.
We formulate the score of a track~$\mathbf{j}=\langle
j^1,\ldots,j^T\rangle$ as
\begin{equation}
  \max_{j^1,\ldots,j^T}
  \sum_{t=1}^Tf(b^t_{j^t})+
  \sum_{t=2}^Tg(b^{t-1}_{j^{t-1}},b^t_{j^t})
  \label{eq:track}
\end{equation}
where we take $g$, the motion coherence, to be a function of the
squared Euclidean distance between the center of $b^{t-1}_{j^{t-1}}$
and the center of $b^t_{j^t}$ projected one frame forward.
While the number of possible tracks is exponential in the number of
frames in the video, Eq.~\ref{eq:track} can be maximized in time
linear in the number of frames and quadratic in the number of
detections per frame using dynamic programming, the \citet{Viterbi1971}
algorithm.

The development of this tracker follows that of \citet{Barbu2012b}
which presents additional details of such a tracker, including an
extension which allows it to generate multiple tracks per object class
by non-maxima suppression.
The tracker employed here has a number of differences from that of
\citet{Barbu2012b}.
While that tracker used the raw detection scores from the
\citet{Felzenszwalb2010b,Felzenszwalb2010a} detector, these scores are
difficult to interpret because the mean score and variance varies by
object class making it difficult to decide whether a detection is strong.
To get around this problem, we pass all detections through a sigmoid
$\frac{1}{1+\exp(-b(t-a))}$ whose center, $a$, is the model threshold
and whose scaling factor $b$, is 2.
This normalizes the score to the range $[0,1]$ and makes scores more
comparable across models.
In addition, the motion coherence score is also passed through a similar
sigmoid, with center 50 and scale $-1/11$.

\section{Word recognition}

Given tracks, we want to decide if a word describes one or more of
those tracks.
This is a generalization of event recognition, generalizing the notion of an
event from verbs to other parts of speech.
To recognize if a word describes a collection of tracks, we extract features
from those tracks and use those features to formulate the semantics of
words.
Word semantics are formulated in terms of finite state machines (FSMs)
which accept one or more tracks.
Fig.~\ref{fig:words} provides an overview of all FSMs used in this paper,
rendered as regular expressions along with their semantics.
This approach is a limiting case of that taken by \citet{Barbu2012a}
which used hidden Markov models (HMMs) to encode the semantics of
words.
In essence, our FSMs are unnormalized HMMs with binary transition
matrices and binary output distributions.
This allows the same recognition mechanism as that used by
\citet{Barbu2012a} to be employed here.

We construct word meaning in two levels.
First, we construct 18 predicates, shown in
Fig.~\ref{fig:predicates}, which accept one or more detections.
We then construct word meanings for our lexicon of 15 words, shown in
Fig.~\ref{fig:words}, as regular expressions which accept tracks and
are composed out of these predicates.
The reason for this two-level construction is to allow for sharing of
low-level features and parameters.
All words share the same predicates which are encoded relative to 9
parameters: $\textbf{far}$, $\textbf{close}$, $\textbf{stationary}$,
$\Delta\textbf{closing}$, $\Delta\textbf{angle}$, $\Delta\textbf{pp}$,
$\Delta\textbf{quickly}$, $\Delta\textbf{slowly}$, and $\textbf{overlap}$.
These parameters are learned from a small number of positive and
negative examples that cover only a small number of words in the
lexicon.
To make predicates independent of the video resolution, detections are first
rescaled relative to a standard resolution of $\text{1280}\times\text{720}$,
otherwise parameters such as $\textbf{far}$ would vary with the
resolution.

Given a regular expression for a word, we can construct a
non-deterministic FSM, with one accepting state, whose
allowable transitions are encoded by a binary transition matrix $h$,
giving score zero to allowed transitions and $-\infty$ to disallowed
transitions, and whose states accept detections which agree with the
predicate $a$, again with the same score of zero or $-\infty$.
With this FSM, we can recognize if a word describes a
track $\langle\hat{\jmath}^1,\ldots,\hat{\jmath}^T\rangle$, by finding
\begin{equation*}
  \max_{k^1,\ldots,k^T}
  \sum_{t=1}^Th(k^t,b^t_{\hat{\jmath}^t})+
  \sum_{t=2}^Ta(k^{t-1},k^t)
\end{equation*}
where $k^1$ through $k^{T-1}$ range over the set of states of the FSM
and $k^T$ is the singleton set containing the accepting state.
If this word describes the track, the score will be zero.
If it does not, the score will be $-\infty$.
The above formulation is trivially generalized to multiple tracks and
is the same as that used by \citet{Barbu2012b}.
We find accepting paths through the lattice of states using dynamic
programming, the Viterbi algorithm.
Note that this method can be applied to encode not just the meaning of
verbs but also of other parts of speech, for example the meaning of
\emph{left-of}.
We will avail ourselves of the ability to encode the meaning of all
parts of speech into a uniform representation in order to build up
the semantics of sentences from the semantics of words.

\begin{figure}[t]
  \centering
  \begin{math}
    \begin{array}{lcl}
      \textsc{far}(a,b) &\define& |a_\textit{cx} - b_\textit{cx}| - \frac{a_\textit{width}}{2} - \frac{b_\textit{width}}{2} > \textbf{far}\\
      \textsc{really-close}(a,b) &\define& |a_\textit{cx} - b_\textit{cx}| - \frac{a_\textit{width}}{2} - \frac{b_\textit{width}}{2} > \frac{\textbf{close}}{2}\\
      \textsc{close}(a,b) &\define& |a_\textit{cx} - b_\textit{cx}| - \frac{a_\textit{width}}{2} - \frac{b_\textit{width}}{2} > \frac{\textbf{close}}{2}\\
      \textsc{stationary}(b) &\define& \text{flow-magnitude}(t) \leq \textbf{stationary}\\
      \textsc{closing}(a,b) &\define& |a_\textit{cx} - b_\textit{cx}| > |\text{project}(a)_\textit{cx} - \text{project}(b)_\textit{cx}| + \Delta\textbf{closing}\\
      \textsc{departing}(a,b) &\define& |a_\textit{cx} - b_\textit{cx}| < |\text{project}(a)_\textit{cx} - \text{project}(b)_\textit{cx}| + \Delta\textbf{closing}\\
      \textsc{moving-direction}(a,b,\alpha) &\define& |\text{flow-orientation}(a) - \alpha|^\circ < \Delta\textbf{angle}\ \wedge{}\\
      &&\ \text{flow-magnitude}(a) > \textbf{stationary}\\
      \textsc{left-of}(a,b) &\define& a_\textit{cx} < b_\textit{cx} + \Delta\textbf{pp}\\
      \textsc{right-of}(a,b) &\define& a_\textit{cx} > b_\textit{cx} + \Delta\textbf{pp}\\
      \textsc{leftward}(a,b) &\define& \textsc{moving-direction}(a,b,0)\\
      \textsc{leftward}(a,b) &\define& \textsc{moving-direction}(a,b,\pi)\\
      \textsc{stationary-but-far}(a,b) &\define& \textsc{far}(a,b) \wedge \textsc{stationary}(a) \wedge \textsc{stationary}(b)\\
      \textsc{stationary-but-close}(a,b) &\define& \textsc{close}(a,b) \wedge \textsc{stationary}(a) \wedge \textsc{stationary}(b)\\
      \textsc{moving-together}(a,b) &\define& |\text{flow-orientation}(a) - \text{flow-orientation}(b)|^\circ < \Delta\textbf{angle} \wedge{}\\
      &&\ \text{flow-magnitude}(a) > \textbf{stationary} \wedge{}\\
      &&\ \text{flow-magnitude}(b) > \textbf{stationary}\\
      \textsc{approaching}(a,b) &\define& \textsc{closing}(a,b) \wedge \textsc{stationary}(b)\\
      \textsc{quickly}(a) &\define& \text{flow-magnitude}(a) > \Delta\textbf{quickly}\\
      \textsc{slowly}(a) &\define& \textbf{stationary} < \text{flow-magnitude}(a) < \Delta\textbf{slowly}\\
      \textsc{overlapping}(a,b) &\define& \frac{a \cap b}{a \cup b} \geq \textbf{overlap}
    \end{array}
  \end{math}
  \caption{Predicates which accept detections, denoted by $a$ and $b$,
    formulated around 9 parameters.
    The function $\text{project}$ projects a detection forward one frame using
    optical flow.
    The functions $\text{flow-orientation}$ and $\text{flow-magnitude}$ compute
    the angle and magnitude of the average optical-flow vector inside a
    detection.
    The function $a_\textit{cx}$ accesses the $x$ coordinate of the center of a
    detection.
    The function $a_\textit{width}$ computes the width of a detection.
    Words are formed as regular expressions over these predicates.}
  \label{fig:predicates}
\end{figure}

\begin{figure}[t]
  \centering
  \begin{math}
    \begin{array}{@{}l@{\hspace{1ex}}c@{\hspace{1ex}}l@{}}
      \texttt{horse}(a) &\define& (a_{\text{object-class}} = ``horse'')^+\\
      \texttt{person}(a) &\define& (a_{\text{object-class}} = ``person'')^+\\
      \texttt{quickly}(a) &\define& \textbf{true}^+\;\textsc{quickly}(a)^{\{3,\}}\;\textbf{true}^+\\
      \texttt{slowly}(a) &\define& \textbf{true}^+\;\textsc{slowly}(a)^{\{3,\}}\;\textbf{true}^+\\
      \texttt{from the left}(a,b) &\define& \textbf{true}^+\;\textsc{left-of}(a,b)^{\{5,\}}\;\textbf{true}^+\\
      \texttt{from the right}(a,b) &\define& \textbf{true}^+\;\textsc{right-of}(a,b)^{\{5,\}}\;\textbf{true}^+\\
      \texttt{leftward}(a) &\define& \textbf{true}^+\;\textsc{leftward}(a)^{\{5,\}}\;\textbf{true}^+\\
      \texttt{rightward}(a) &\define& \textbf{true}^+\;\textsc{rightward}(a)^{\{5,\}}\;\textbf{true}^+\\
      \texttt{to the left of}(a,b) &\define& \textbf{true}^+\;\textsc{left-of}(a,b)^{\{3,\}}\;\textbf{true}^+\\
      \texttt{to the right of}(a,b) &\define& \textbf{true}^+\;\textsc{right-of}(a,b)^{\{3,\}}\;\textbf{true}^+\\
      \texttt{towards}(a,b) &\define&
      \begin{array}[t]{@{}l@{}}
        \textsc{stationary-but-far}(a,b)^+\;\textsc{approaching}(a,b)^{\{3,\}}\\
        \textsc{stationary-but-close}(a,b)^+
      \end{array}\\
      \texttt{away from}(a,b) &\define&
      \begin{array}[t]{@{}l@{}}
        \textsc{stationary-but-close}(a,b)^+\;\textsc{departing}(a,b)^{\{3,\}}\\
        \textsc{stationary-but-far}(a,b)^+
      \end{array}\\
      \texttt{ride}(a,b) &\define& \textbf{true}^+\;(\textsc{moving-together}(a,b)\wedge\textsc{overlapping}(a,b))^{\{5,\}}\;\textbf{true}^+\\
      \texttt{lead}(a,b) &\define& \textbf{true}^+\;\left(
        \begin{array}{l}
          \neg\textsc{really-close}(a,b)\wedge{}\\
          \textsc{moving-together}(a,b)\wedge{}\\
          \left(\begin{array}{@{}l@{}}
              (\textsc{left-of}(a,b)\wedge\textsc{leftward}(a))\vee{}\\
              (\textsc{right-of}(a,b)\wedge\textsc{rightward}(a))
            \end{array}\right)
        \end{array}
      \right)^{\{5,\}}\;\textbf{true}^+\\
      \texttt{approach}(a,b) &\define& \textbf{true}^+\;\textsc{approaching}(a,b)^{\{5,\}}\;\textbf{true}^+
    \end{array}
  \end{math}
  \caption{Regular expressions which encode the meanings of each of
    the 15 words or lexicalized phrases in the lexicon as regular
    expressions composed of the predicates shown in
    Fig.~\ref{fig:predicates}.
    We use an extended regular expression
    syntax where an exponent of $+$ allows a predicate to hold for one or more
    frames and exponent of $\{t,\}$ allows a predicate to hold for $t$ or
    more frames.}
  \label{fig:words}
\end{figure}

\section{Sentence tracker}

Our ultimate goal is to search for videos given a natural-language
query in the form of a sentence.
The framework developed so far falls short of supporting this goal in
two ways.
First, as we attempt to recognize multiple words that constrain a
single track, it becomes unlikely that the tracker will happen to produce an
optimal track which satisfies all the desired predicates.
For example, we want a person that is both \emph{running} and doing so
\emph{leftward}.
Second, a sentence is not a conjunction of words, even though a word
is represented here as a conjunction of features, so a new mechanism
is required to faithfully encode the semantics of a sentence.
Intuitively, we need a way to encode the mutual dependence in the
sentence \emph{The tall person rode the horse} so that the person is
tall, not the horse, and the person is riding the horse, not
vice versa.

We address the first point by biasing the tracker to produce tracks
which agree with the predicates that are being enforced.
This may result in the tracker producing tracks which have to consist
of lower-scoring detections, which decreases the probability that these
tracks correspond to real objects in the video,
This is not a concern as we will present the users with results ranked by their
tracker score.
In essence, we pay a penalty for forcing a track to agree with the enforced
predicates and the ultimate rank order is influenced by this penalty.
The computational mechanism that enables this exists by virtue of the fact that
our tracker and word recognizer have the same internal representation and
algorithm, namely, each finds optimal paths through a lattice of detections and
states, respectively, and each weights the links in that lattice by a score,
the motion coherence and state-transition score, respectively.
We simultaneously find the optimal, highest-scoring, track
${j^1},\ldots,{j^T}$ and state sequence ${k^1},\ldots,{k^T}$ as
\begin{equation*}
  \max_{{j^1},\ldots,{j^T}}
  \max_{{k^1},\ldots,{k^T}}
  \displaystyle\sum_{t=1}^T f(b^t_{{j^t}})
  +\displaystyle\sum_{t=2}^T g(b^{t-1}_{{j^{t-1}}},b^t_{{j^t}})
  +\displaystyle\sum_{t=1}^T h(k^t,b^t_{j^t})
  +\displaystyle\sum_{t=2}^T a(k^{t-1},k^t)
\end{equation*}
which ensures that, unless the state sequence for the word FSM leads to
an accepting state, the resulting score will be $-\infty$ and thereby
constrains the tracks to depict the word.
Intuitively, we have two lattices, a tracker lattice and a word-recognizer
lattice, and we find the optimal path, again with the
Viterbi algorithm, through a cross-product lattice.

The above handles only a single word, but given a sentential query we
want to encode its semantics in terms of multiple words and multiple
trackers.
We parse an input sentence with a grammar, shown in
Fig.~\ref{fig:grammar}, and extract the number of participants and
the track-to-role mapping.
Each sentence has a number of thematic roles that must be filled by
participants in order for the sentence to be syntactically valid.
For example, in the sentence \emph{The person rode the horse quickly
  away from the other horse}, there are three participants, one
person and two horses, and each of the three participants plays a
different role in the sentence, \emph{agent}, \emph{patient}, and
\emph{goal}.
Each word in this sentence refers to a subset of these three different
participants, as shown in Fig.~\ref{fig:sentence}, and words that
refer to multiple participants, such as \emph{ride}, must be assigned
participants in the correct order to ensure that we encode \emph{The
  person rode the horse} rather than \emph{The horse rode the person}.
We use a custom natural-language parser which takes as input a
grammar, along with the arity and thematic roles of each word, and
computes a track-to-role mapping: which participants fill which roles
in which words.
We employ the same mechanism as described above for simultaneous word
recognition and tracking, except that we instantiate one tracker for
each participant and one word recognizer for each word.
The thematic roles, $\theta^n_w$, map the $n$th role in a word $w$ to
a tracker.
Fig.~\ref{fig:sentence} displays an overview of this mapping for a
sample sentence.
Trackers are shown in red, word recognizers are shown in blue, and the
track-to-role mapping is shown using the arrows.
Given a sentential query that has $W$ words, $L$ participants, and
track-to-role mapping $\theta^n_w$, we find a collection of optimal tracks
$\langle j^1_1,\ldots,j^T_1\rangle\cdots\langle j^1_L,\ldots,j^T_L\rangle$,
one for each participant, and accepting state sequences
$\langle k^1_1,\ldots,k^T_1\rangle\cdots\langle k^1_W,\ldots,k^T_W\rangle$,
one for each word, as
\begin{equation*}
  \max_{\substack{j^1_1,\ldots,j^T_1\\j^1_L,\ldots,j^T_L}}
  \max_{\substack{k^1_1,\ldots,k^T_1\\k^1_W,\ldots,k^T_W}}
  \sum_{l=1}^L
  \sum_{t=1}^Tf(b^t_{j^t_l})+
  \sum_{t=2}^Tg(b^{t-1}_{j^{t-1}_l},b^t_{j^t_l})+
  \sum_{w=1}^W
  \sum_{t=1}^Th_w(k^t_w,b^t_{j^t_{\theta^1_w}},b^t_{j^t_{\theta^2_w}})+
  \sum_{t=2}^Ta_w(k^{t-1}_w,k^t_w)
\end{equation*}
where $a_w$ and $h_w$ are the transition matrices and predicates for word
$w$, $b^t_{j^t_l}$ is a detection in the $t$th frame of the $l$th
track, and $b^t_{j^t_{\theta^n_w}}$ connects a participant that fills
the $n$th role in word $w$ with the detections of its tracker.
This equation maximizes the tracker score for each tracker
corresponding to each participant, and ensures that each word has a
sequence of accepting states, if such a sequence exists, otherwise the
entire sentence-tracker score will be $-\infty$.
In essence, we are taking cross products of tracker lattices and word
lattices while ensuring that the sequence of cross products agrees
with our track-to-role mapping and finding the optimal path through
the resulting lattice.
This allows us to employ the same computational mechanism, the Viterbi
algorithm, to find this optimal node sequence.
The resulting tracks will satisfy the semantics of the input sentence,
even if this means paying a penalty by having to choose lower-scoring
detections.

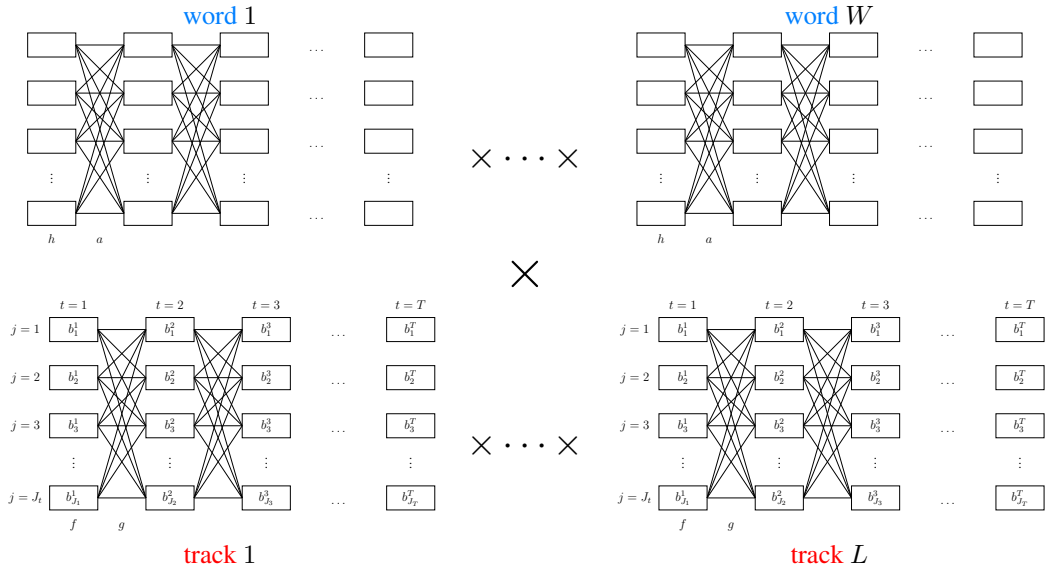
\begin{figure}[t]
  \begin{center}
    \begin{tabular}{ccc}
      {\lightBlue word} $1$&&{\lightBlue word} $W$\\
      \scalebox{0.4}{\begin{tikzpicture}[y=-1cm]

\draw[black] (3.8,4.2) -- (5.4,4.2);
\draw[black] (3.8,2.6) -- (5.4,2.6);
\draw[black] (3.8,5.8) -- (5.4,5.8);
\draw[black] (3.8,8.2) -- (5.4,8.2);
\draw[black] (3.8,2.6) -- (5.4,4.2);
\draw[black] (3.8,2.6) -- (5.4,5.8);
\draw[black] (3.8,2.6) -- (5.4,8.2);
\draw[black] (3.8,4.2) -- (5.4,2.6);
\draw[black] (3.8,4.2) -- (5.4,5.8);
\draw[black] (3.8,4.2) -- (5.4,8.2);
\draw[black] (3.8,5.8) -- (5.4,2.6);
\draw[black] (3.8,5.8) -- (5.4,4.2);
\draw[black] (3.8,5.8) -- (5.4,8.2);
\draw[black] (3.8,8.2) -- (5.4,2.6);
\draw[black] (3.8,8.2) -- (5.4,4.2);
\draw[black] (3.8,8.2) -- (5.4,5.8);
\draw[black] (7,4.2) -- (8.6,4.2);
\draw[black] (7,2.6) -- (8.6,2.6);
\draw[black] (7,5.8) -- (8.6,5.8);
\draw[black] (7,8.2) -- (8.6,8.2);
\draw[black] (7,2.6) -- (8.6,4.2);
\draw[black] (7,2.6) -- (8.6,5.8);
\draw[black] (7,2.6) -- (8.6,8.2);
\draw[black] (7,4.2) -- (8.6,2.6);
\draw[black] (7,4.2) -- (8.6,5.8);
\draw[black] (7,4.2) -- (8.6,8.2);
\draw[black] (7,5.8) -- (8.6,2.6);
\draw[black] (7,5.8) -- (8.6,4.2);
\draw[black] (7,5.8) -- (8.6,8.2);
\draw[black] (7,8.2) -- (8.6,2.6);
\draw[black] (7,8.2) -- (8.6,4.2);
\draw[black] (7,8.2) -- (8.6,5.8);
\draw[black] (2.2,2.2) rectangle (3.8,3);
\draw[black] (2.2,3.8) rectangle (3.8,4.6);
\draw[black] (2.2,5.4) rectangle (3.8,6.2);
\draw[black] (2.2,7.8) rectangle (3.8,8.6);
\path (3,7.2) node[text=black,anchor=base] {\large{}$\vdots$};
\draw[black] (5.4,2.2) rectangle (7,3);
\draw[black] (5.4,3.8) rectangle (7,4.6);
\draw[black] (5.4,5.4) rectangle (7,6.2);
\draw[black] (5.4,7.8) rectangle (7,8.6);
\path (6.2,7.2) node[text=black,anchor=base] {\large{}$\vdots$};
\draw[black] (8.6,2.2) rectangle (10.2,3);
\draw[black] (8.6,3.8) rectangle (10.2,4.6);
\draw[black] (8.6,5.4) rectangle (10.2,6.2);
\draw[black] (8.6,7.8) rectangle (10.2,8.6);
\path (9.4,7.2) node[text=black,anchor=base] {\large{}$\vdots$};
\draw[black] (13.4,2.2) rectangle (15,3);
\draw[black] (13.4,3.8) rectangle (15,4.6);
\draw[black] (13.4,5.4) rectangle (15,6.2);
\draw[black] (13.4,7.8) rectangle (15,8.6);
\path (14.2,7.2) node[text=black,anchor=base] {\large{}$\vdots$};
\path (11.8,2.8) node[text=black,anchor=base] {\large{}$\ldots$};
\path (11.8,4.4) node[text=black,anchor=base] {\large{}$\ldots$};
\path (11.8,6) node[text=black,anchor=base] {\large{}$\ldots$};
\path (11.8,8.4) node[text=black,anchor=base] {\large{}$\ldots$};
\path (3,9.2) node[text=black,anchor=base] {\large{}$h$};
\path (4.6,9.2) node[text=black,anchor=base] {\large{}$a$};

\end{tikzpicture}%
}&
      \raisebox{30pt}{{\Large $\times\cdots\times$}}&
      \scalebox{0.4}{\begin{tikzpicture}[y=-1cm]

\draw[black] (3.8,4.2) -- (5.4,4.2);
\draw[black] (3.8,2.6) -- (5.4,2.6);
\draw[black] (3.8,5.8) -- (5.4,5.8);
\draw[black] (3.8,8.2) -- (5.4,8.2);
\draw[black] (3.8,2.6) -- (5.4,4.2);
\draw[black] (3.8,2.6) -- (5.4,5.8);
\draw[black] (3.8,2.6) -- (5.4,8.2);
\draw[black] (3.8,4.2) -- (5.4,2.6);
\draw[black] (3.8,4.2) -- (5.4,5.8);
\draw[black] (3.8,4.2) -- (5.4,8.2);
\draw[black] (3.8,5.8) -- (5.4,2.6);
\draw[black] (3.8,5.8) -- (5.4,4.2);
\draw[black] (3.8,5.8) -- (5.4,8.2);
\draw[black] (3.8,8.2) -- (5.4,2.6);
\draw[black] (3.8,8.2) -- (5.4,4.2);
\draw[black] (3.8,8.2) -- (5.4,5.8);
\draw[black] (7,4.2) -- (8.6,4.2);
\draw[black] (7,2.6) -- (8.6,2.6);
\draw[black] (7,5.8) -- (8.6,5.8);
\draw[black] (7,8.2) -- (8.6,8.2);
\draw[black] (7,2.6) -- (8.6,4.2);
\draw[black] (7,2.6) -- (8.6,5.8);
\draw[black] (7,2.6) -- (8.6,8.2);
\draw[black] (7,4.2) -- (8.6,2.6);
\draw[black] (7,4.2) -- (8.6,5.8);
\draw[black] (7,4.2) -- (8.6,8.2);
\draw[black] (7,5.8) -- (8.6,2.6);
\draw[black] (7,5.8) -- (8.6,4.2);
\draw[black] (7,5.8) -- (8.6,8.2);
\draw[black] (7,8.2) -- (8.6,2.6);
\draw[black] (7,8.2) -- (8.6,4.2);
\draw[black] (7,8.2) -- (8.6,5.8);
\draw[black] (2.2,2.2) rectangle (3.8,3);
\draw[black] (2.2,3.8) rectangle (3.8,4.6);
\draw[black] (2.2,5.4) rectangle (3.8,6.2);
\draw[black] (2.2,7.8) rectangle (3.8,8.6);
\path (3,7.2) node[text=black,anchor=base] {\large{}$\vdots$};
\draw[black] (5.4,2.2) rectangle (7,3);
\draw[black] (5.4,3.8) rectangle (7,4.6);
\draw[black] (5.4,5.4) rectangle (7,6.2);
\draw[black] (5.4,7.8) rectangle (7,8.6);
\path (6.2,7.2) node[text=black,anchor=base] {\large{}$\vdots$};
\draw[black] (8.6,2.2) rectangle (10.2,3);
\draw[black] (8.6,3.8) rectangle (10.2,4.6);
\draw[black] (8.6,5.4) rectangle (10.2,6.2);
\draw[black] (8.6,7.8) rectangle (10.2,8.6);
\path (9.4,7.2) node[text=black,anchor=base] {\large{}$\vdots$};
\draw[black] (13.4,2.2) rectangle (15,3);
\draw[black] (13.4,3.8) rectangle (15,4.6);
\draw[black] (13.4,5.4) rectangle (15,6.2);
\draw[black] (13.4,7.8) rectangle (15,8.6);
\path (14.2,7.2) node[text=black,anchor=base] {\large{}$\vdots$};
\path (11.8,2.8) node[text=black,anchor=base] {\large{}$\ldots$};
\path (11.8,4.4) node[text=black,anchor=base] {\large{}$\ldots$};
\path (11.8,6) node[text=black,anchor=base] {\large{}$\ldots$};
\path (11.8,8.4) node[text=black,anchor=base] {\large{}$\ldots$};
\path (3,9.2) node[text=black,anchor=base] {\large{}$h$};
\path (4.6,9.2) node[text=black,anchor=base] {\large{}$a$};

\end{tikzpicture}%
}\\
      &{\huge $\times$}&\\
      \scalebox{0.4}{\begin{tikzpicture}[y=-1cm]

\draw[black] (3.8,4.2) -- (5.4,4.2);
\draw[black] (3.8,2.6) -- (5.4,2.6);
\draw[black] (3.8,5.8) -- (5.4,5.8);
\draw[black] (3.8,8.2) -- (5.4,8.2);
\draw[black] (3.8,2.6) -- (5.4,4.2);
\draw[black] (3.8,2.6) -- (5.4,5.8);
\draw[black] (3.8,2.6) -- (5.4,8.2);
\draw[black] (3.8,4.2) -- (5.4,2.6);
\draw[black] (3.8,4.2) -- (5.4,5.8);
\draw[black] (3.8,4.2) -- (5.4,8.2);
\draw[black] (3.8,5.8) -- (5.4,2.6);
\draw[black] (3.8,5.8) -- (5.4,4.2);
\draw[black] (3.8,5.8) -- (5.4,8.2);
\draw[black] (3.8,8.2) -- (5.4,2.6);
\draw[black] (3.8,8.2) -- (5.4,4.2);
\draw[black] (3.8,8.2) -- (5.4,5.8);
\draw[black] (7,4.2) -- (8.6,4.2);
\draw[black] (7,2.6) -- (8.6,2.6);
\draw[black] (7,5.8) -- (8.6,5.8);
\draw[black] (7,8.2) -- (8.6,8.2);
\draw[black] (7,2.6) -- (8.6,4.2);
\draw[black] (7,2.6) -- (8.6,5.8);
\draw[black] (7,2.6) -- (8.6,8.2);
\draw[black] (7,4.2) -- (8.6,2.6);
\draw[black] (7,4.2) -- (8.6,5.8);
\draw[black] (7,4.2) -- (8.6,8.2);
\draw[black] (7,5.8) -- (8.6,2.6);
\draw[black] (7,5.8) -- (8.6,4.2);
\draw[black] (7,5.8) -- (8.6,8.2);
\draw[black] (7,8.2) -- (8.6,2.6);
\draw[black] (7,8.2) -- (8.6,4.2);
\draw[black] (7,8.2) -- (8.6,5.8);
\draw[black] (2.2,2.2) rectangle (3.8,3);
\draw[black] (2.2,3.8) rectangle (3.8,4.6);
\draw[black] (2.2,5.4) rectangle (3.8,6.2);
\draw[black] (2.2,7.8) rectangle (3.8,8.6);
\path (3,7.2) node[text=black,anchor=base] {\large{}$\vdots$};
\draw[black] (5.4,2.2) rectangle (7,3);
\draw[black] (5.4,3.8) rectangle (7,4.6);
\draw[black] (5.4,5.4) rectangle (7,6.2);
\draw[black] (5.4,7.8) rectangle (7,8.6);
\path (6.2,7.2) node[text=black,anchor=base] {\large{}$\vdots$};
\draw[black] (8.6,2.2) rectangle (10.2,3);
\draw[black] (8.6,3.8) rectangle (10.2,4.6);
\draw[black] (8.6,5.4) rectangle (10.2,6.2);
\draw[black] (8.6,7.8) rectangle (10.2,8.6);
\path (9.4,7.2) node[text=black,anchor=base] {\large{}$\vdots$};
\draw[black] (13.4,2.2) rectangle (15,3);
\draw[black] (13.4,3.8) rectangle (15,4.6);
\draw[black] (13.4,5.4) rectangle (15,6.2);
\draw[black] (13.4,7.8) rectangle (15,8.6);
\path (14.2,7.2) node[text=black,anchor=base] {\large{}$\vdots$};
\path (11.8,2.8) node[text=black,anchor=base] {\large{}$\ldots$};
\path (11.8,4.4) node[text=black,anchor=base] {\large{}$\ldots$};
\path (11.8,6) node[text=black,anchor=base] {\large{}$\ldots$};
\path (11.8,8.4) node[text=black,anchor=base] {\large{}$\ldots$};
\path (3,9.2) node[text=black,anchor=base] {\large{}$f$};
\path (4.6,9.2) node[text=black,anchor=base] {\large{}$g$};
\path (3,1.9) node[text=black,anchor=base] {\large{}$t=1$};
\path (6.2,1.9) node[text=black,anchor=base] {\large{}$t=2$};
\path (9.4,1.9) node[text=black,anchor=base] {\large{}$t=3$};
\path (14.2,1.9) node[text=black,anchor=base] {\large{}$t=T$};
\path (1.4,2.7) node[text=black,anchor=base] {\large{}$j=1$};
\path (1.4,5.9) node[text=black,anchor=base] {\large{}$j=3$};
\path (1.4,4.3) node[text=black,anchor=base] {\large{}$j=2$};
\path (1.4,8.3) node[text=black,anchor=base] {\large{}$j=J_t$};
\path (3,2.7) node[text=black,anchor=base] {\large{}$b^1_1$};
\path (3,4.3) node[text=black,anchor=base] {\large{}$b^1_2$};
\path (3,5.9) node[text=black,anchor=base] {\large{}$b^1_3$};
\path (3,8.3) node[text=black,anchor=base] {\large{}$b^1_{J_1}$};
\path (6.2,2.7) node[text=black,anchor=base] {\large{}$b^2_1$};
\path (6.2,4.3) node[text=black,anchor=base] {\large{}$b^2_2$};
\path (6.2,5.9) node[text=black,anchor=base] {\large{}$b^2_3$};
\path (6.2,8.3) node[text=black,anchor=base] {\large{}$b^2_{J_2}$};
\path (9.4,2.7) node[text=black,anchor=base] {\large{}$b^3_1$};
\path (9.4,4.3) node[text=black,anchor=base] {\large{}$b^3_2$};
\path (9.4,5.9) node[text=black,anchor=base] {\large{}$b^3_3$};
\path (9.4,8.3) node[text=black,anchor=base] {\large{}$b^3_{J_3}$};
\path (14.2,2.7) node[text=black,anchor=base] {\large{}$b^T_1$};
\path (14.2,4.3) node[text=black,anchor=base] {\large{}$b^T_2$};
\path (14.2,5.9) node[text=black,anchor=base] {\large{}$b^T_3$};
\path (14.2,8.3) node[text=black,anchor=base] {\large{}$b^T_{J_T}$};

\end{tikzpicture}%
}&
      \raisebox{30pt}{{\Large $\times\cdots\times$}}&
      \scalebox{0.4}{\begin{tikzpicture}[y=-1cm]

\draw[black] (3.8,4.2) -- (5.4,4.2);
\draw[black] (3.8,2.6) -- (5.4,2.6);
\draw[black] (3.8,5.8) -- (5.4,5.8);
\draw[black] (3.8,8.2) -- (5.4,8.2);
\draw[black] (3.8,2.6) -- (5.4,4.2);
\draw[black] (3.8,2.6) -- (5.4,5.8);
\draw[black] (3.8,2.6) -- (5.4,8.2);
\draw[black] (3.8,4.2) -- (5.4,2.6);
\draw[black] (3.8,4.2) -- (5.4,5.8);
\draw[black] (3.8,4.2) -- (5.4,8.2);
\draw[black] (3.8,5.8) -- (5.4,2.6);
\draw[black] (3.8,5.8) -- (5.4,4.2);
\draw[black] (3.8,5.8) -- (5.4,8.2);
\draw[black] (3.8,8.2) -- (5.4,2.6);
\draw[black] (3.8,8.2) -- (5.4,4.2);
\draw[black] (3.8,8.2) -- (5.4,5.8);
\draw[black] (7,4.2) -- (8.6,4.2);
\draw[black] (7,2.6) -- (8.6,2.6);
\draw[black] (7,5.8) -- (8.6,5.8);
\draw[black] (7,8.2) -- (8.6,8.2);
\draw[black] (7,2.6) -- (8.6,4.2);
\draw[black] (7,2.6) -- (8.6,5.8);
\draw[black] (7,2.6) -- (8.6,8.2);
\draw[black] (7,4.2) -- (8.6,2.6);
\draw[black] (7,4.2) -- (8.6,5.8);
\draw[black] (7,4.2) -- (8.6,8.2);
\draw[black] (7,5.8) -- (8.6,2.6);
\draw[black] (7,5.8) -- (8.6,4.2);
\draw[black] (7,5.8) -- (8.6,8.2);
\draw[black] (7,8.2) -- (8.6,2.6);
\draw[black] (7,8.2) -- (8.6,4.2);
\draw[black] (7,8.2) -- (8.6,5.8);
\draw[black] (2.2,2.2) rectangle (3.8,3);
\draw[black] (2.2,3.8) rectangle (3.8,4.6);
\draw[black] (2.2,5.4) rectangle (3.8,6.2);
\draw[black] (2.2,7.8) rectangle (3.8,8.6);
\path (3,7.2) node[text=black,anchor=base] {\large{}$\vdots$};
\draw[black] (5.4,2.2) rectangle (7,3);
\draw[black] (5.4,3.8) rectangle (7,4.6);
\draw[black] (5.4,5.4) rectangle (7,6.2);
\draw[black] (5.4,7.8) rectangle (7,8.6);
\path (6.2,7.2) node[text=black,anchor=base] {\large{}$\vdots$};
\draw[black] (8.6,2.2) rectangle (10.2,3);
\draw[black] (8.6,3.8) rectangle (10.2,4.6);
\draw[black] (8.6,5.4) rectangle (10.2,6.2);
\draw[black] (8.6,7.8) rectangle (10.2,8.6);
\path (9.4,7.2) node[text=black,anchor=base] {\large{}$\vdots$};
\draw[black] (13.4,2.2) rectangle (15,3);
\draw[black] (13.4,3.8) rectangle (15,4.6);
\draw[black] (13.4,5.4) rectangle (15,6.2);
\draw[black] (13.4,7.8) rectangle (15,8.6);
\path (14.2,7.2) node[text=black,anchor=base] {\large{}$\vdots$};
\path (11.8,2.8) node[text=black,anchor=base] {\large{}$\ldots$};
\path (11.8,4.4) node[text=black,anchor=base] {\large{}$\ldots$};
\path (11.8,6) node[text=black,anchor=base] {\large{}$\ldots$};
\path (11.8,8.4) node[text=black,anchor=base] {\large{}$\ldots$};
\path (3,9.2) node[text=black,anchor=base] {\large{}$f$};
\path (4.6,9.2) node[text=black,anchor=base] {\large{}$g$};
\path (3,1.9) node[text=black,anchor=base] {\large{}$t=1$};
\path (6.2,1.9) node[text=black,anchor=base] {\large{}$t=2$};
\path (9.4,1.9) node[text=black,anchor=base] {\large{}$t=3$};
\path (14.2,1.9) node[text=black,anchor=base] {\large{}$t=T$};
\path (1.4,2.7) node[text=black,anchor=base] {\large{}$j=1$};
\path (1.4,5.9) node[text=black,anchor=base] {\large{}$j=3$};
\path (1.4,4.3) node[text=black,anchor=base] {\large{}$j=2$};
\path (1.4,8.3) node[text=black,anchor=base] {\large{}$j=J_t$};
\path (3,2.7) node[text=black,anchor=base] {\large{}$b^1_1$};
\path (3,4.3) node[text=black,anchor=base] {\large{}$b^1_2$};
\path (3,5.9) node[text=black,anchor=base] {\large{}$b^1_3$};
\path (3,8.3) node[text=black,anchor=base] {\large{}$b^1_{J_1}$};
\path (6.2,2.7) node[text=black,anchor=base] {\large{}$b^2_1$};
\path (6.2,4.3) node[text=black,anchor=base] {\large{}$b^2_2$};
\path (6.2,5.9) node[text=black,anchor=base] {\large{}$b^2_3$};
\path (6.2,8.3) node[text=black,anchor=base] {\large{}$b^2_{J_2}$};
\path (9.4,2.7) node[text=black,anchor=base] {\large{}$b^3_1$};
\path (9.4,4.3) node[text=black,anchor=base] {\large{}$b^3_2$};
\path (9.4,5.9) node[text=black,anchor=base] {\large{}$b^3_3$};
\path (9.4,8.3) node[text=black,anchor=base] {\large{}$b^3_{J_3}$};
\path (14.2,2.7) node[text=black,anchor=base] {\large{}$b^T_1$};
\path (14.2,4.3) node[text=black,anchor=base] {\large{}$b^T_2$};
\path (14.2,5.9) node[text=black,anchor=base] {\large{}$b^T_3$};
\path (14.2,8.3) node[text=black,anchor=base] {\large{}$b^T_{J_T}$};

\end{tikzpicture}%
}\\
      {\red track} $1$&&{\red track} $L$\\
    \end{tabular}
  \end{center}
  \caption{
    Tracker lattices are used to produce tracks for each
    the object.
    Word lattices constructed from word FSMs recognize one or more tracks.
    We take the cross product of multiple tracker lattices and word
    lattices to simultaneously track objects and recognize words.
    By construction, this ensures that the resulting tracks
    are described by the desired words.}
  \label{fig:sentence-tracker}
\end{figure}

\begin{figure}[t]
  \begin{center}
    \begin{tikzpicture}[scale=7]
      \node[] (the) at (0.1,0.9) {\lightBlue The\strut};
      \node[] (tall) at ($(the.east)+(0.08em,0)$) {\lightBlue tall\strut};
      \node[] (person) at ($(tall.east)+(0.18em,0)$) {\lightBlue person\strut};
      \node[] (rode) at ($(person.east)+(0.1em,0)$) {\lightBlue rode\strut};
      \node[] (thetwo) at ($(rode.east)+(0.07em,0)$) {\lightBlue the\strut};
      \node[] (horse) at ($(thetwo.east)+(0.14em,0)$) {\lightBlue horse\strut};
      \node[] (quickly) at ($(horse.east)+(0.22em,0)$) {\lightBlue quickly\strut};
      \node[] (leftward) at ($(quickly.east)+(0.22em,0)$) {\lightBlue leftward\strut};
      \node[] (away) at ($(leftward.east)+(0.13em,0)$) {\lightBlue away\strut};
      \node[] (from) at ($(away.east)+(0.12em,0)$) {\lightBlue from\strut};
      \node[] (thethree) at ($(from.east)+(0.08em,0)$) {\lightBlue the\strut};
      \node[] (other) at ($(thethree.east)+(0.14em,0)$) {\lightBlue other\strut};
      \node[] (horsetwo) at ($(other.east)+(0.14em,0)$) {\lightBlue horse\strut};
      \node[] (agent) at (0.4, 0.5) {{\red agent-track}};
      \node[] (patient) at (0.8, 0.5) {{\red patient-track}};
      \draw[-] ($(person.south)$) -- ($(agent.north)$);
      \draw[-] ($(horse.south)$) -- ($(patient.north)$);
      \node[] (source) at (1.2, 0.5) {{\red source-track}};
      \draw[-] ($(horsetwo.south)$) -- ($(source.north)$);
      \draw[-] ($(rode.south)$) -- ($(agent.north)$);
      \draw[-,dashed] ($(rode.south)$) -- ($(patient.north)$);
      \draw[-] ($(tall.south)$) -- ($(agent.north)$);
      \draw[-] ($(quickly.south)$) -- ($(patient.north)$);
      \draw[-] ($(leftward.south)$) -- ($(patient.north)$);
      \draw[-] ($(away.south east)$) -- ($(patient.north)$);
      \draw[-,dashed] ($(away.south east)$) -- ($(source.north)$);
    \end{tikzpicture}
  \end{center}
  \caption{
    The order of cross products required to encode the meaning of
    a sentence is not arbitrary and is shown here by the arrows
    connecting each word to each tracker, shown in red.
    The number of tracks is determined by parsing the sentence.
    The lattices for words or lexicalized phrases such as \emph{away
      from} are cross producted with the tracks that those words
    refer to.
    The dashed line indicate that the order of the cross products is
    essential for words which have more than one role, in other words
    \emph{rode} is not symmetric.}
  \label{fig:sentence}
\end{figure}
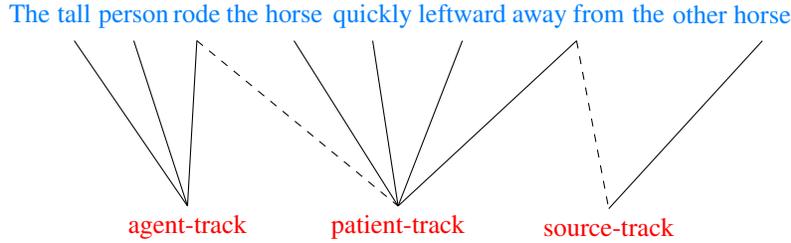

\begin{figure}[t]
  \begin{center}
    \begin{footnotesize}
      \begin{tabular}{@{}l@{\hspace*{10pt}}c@{\hspace*{10pt}}l@{\hspace{40pt}}l@{\hspace*{10pt}}c@{\hspace*{10pt}}l@{}}
        S & $\rightarrow$ & NP VP &
        NP & $\rightarrow$ & D N [PP]\\
        D & $\rightarrow$ & \emph{the} &
        N & $\rightarrow$ & \emph{person} $|$ \emph{horse}\\
        PP & $\rightarrow$ & P NP &
        P & $\rightarrow$ & \emph{to the left of} $|$ \emph{to the right of}\\
        VP & $\rightarrow$ & V NP [Adv] [$\text{PP}_{\text{M}}$] &
        V & $\rightarrow$ & \emph{lead} $|$ \emph{rode} $|$ \emph{approached}\\
        Adv & $\rightarrow$ & \emph{quickly} $|$ \emph{slowly} &
        $\text{PP}_{\text{M}}$ & $\rightarrow$ & $\text{P}_{\text{M}}$ NP $|$ \emph{from the left} $|$ \emph{from the right}\\
        $\text{P}_{\text{M}}$ & $\rightarrow$ & \emph{towards} $|$ \emph{away from}
      \end{tabular}
    \end{footnotesize}
  \end{center}
  \caption{The grammar for sentential queries used in the experiment section.}
  \label{fig:grammar}
\end{figure}

\section{Results}

We have so far developed a system which scores a video-sentence pair
telling us how well a video depicts a sentence.
Given a sentential query, we run the sentence tracker on every video in
a corpus and return all results ranked by their scores.
The better the score the more confident we are that the resulting
tracks correspond to real objects in the video while the sentence
tracker itself ensures that all tracks produced satisfy the sentential
query.
To save on redundant computation, we cache the object-detector results
for each video as the detection scores are independent of the
sentential query.

To demonstrate this approach to video search, we ran sentential queries
over a corpus of 10 Hollywood westerns:
Black Beauty (Warner Brothers, 1994),
The Black Stallion (MGM, 1979),
Blazing Saddles (Warner Brothers, 1974),
Easy Rider (Columbia Pictures, 1969),
The Good the Bad and the Ugly (Columbia Pictures, 1966),
Hidalgo (Touchstone Pictures, 2004),
National Velvet (MGM, 1944),
Once Upon a Time in Mexico (Columbia Pictures, 2003),
Seabiscuit (Universal Pictures, 2003), and
Unforgiven (Warner Brothers, 1992).
In total, this video corpus has 1187 minutes of video, roughly 20
hours.
We temporally downsample all videos to 6 frames per second but keep
their original spatial resolutions which varied from
$\text{336}\times\text{256}$ pixels to $\text{1280}\times\text{544}$ pixels
with a mean resolution of $\text{659.2}\times\text{332.8}$ pixels.
We split these videos into 37187 clips, each clip being 18 frames (3 seconds)
long, which overlaps the previous clip by 6 frames.
This overlap ensures that actions that might otherwise occur on clip
boundaries will also occur as part of a clip.
While there is prior work on shot segmentation \citep{cooper2007video}
we do not employ it for two reasons.
First, it complicates the system and provides an avenue for additional
failure modes.
Second, the approach taken here is able to find an event inside
a longer video with multiple events.
The only reason why we split up the videos into clips is to return
multiple such events.

We adopt the grammar from Fig.~\ref{fig:grammar} which allows for
sentences that describe people interacting with horses, hence our
choice of genre for the video corpus, namely westerns.
A requirement for determining whether a video depicts a sentence and the degree
to which it depicts that sentence is to detect the objects that might
fill roles in that sentence.
Previous work has shown that people and horses are among the easiest-to-detect
objects, although the performance of object detectors, even for these
classes, remains extremely low.
To ensure that we are not testing on the training data, we employ
previously-trained object models that have not been trained on these
videos but have instead been trained on the PASCAL VOC Challenge
\citep{Everingham10}.
We also require settings for the 9 parameters, shown in
Fig.~\ref{fig:predicates}, which are required to produce the
predicates which encode the semantics of the words in this grammar.
We train all 9 parameters simultaneously on only 3 positive examples
and 3 negative examples.
Note that these training examples cover only a subset of the words in the
grammar but are sufficient to define the semantics of all words because this
word subset touches upon all the underlying parameters.
Training proceeds by exhaustively searching a small uniform grid, with
between 3 and 10 steps per dimension, of all nine parameter settings
to find a combination which best classifies all 6 training samples
which are then removed from the test set.
\citet{yu2013grounded} present a related alternative strategy for
training the parameters of a lexicon of words given a video corpus.

\begin{figure}
  \begin{tabular}{ccc}
    \includegraphics[width=0.3\textwidth]{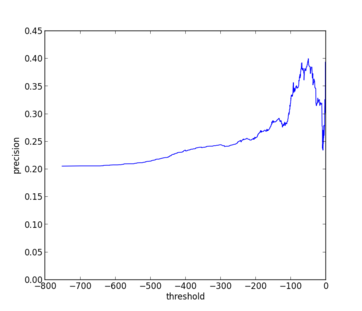}&
    \includegraphics[width=0.3\textwidth]{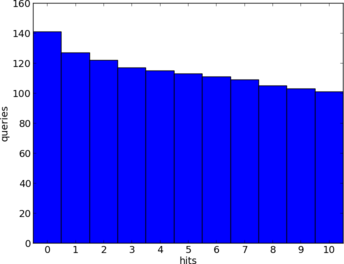}&
    \includegraphics[width=0.3\textwidth]{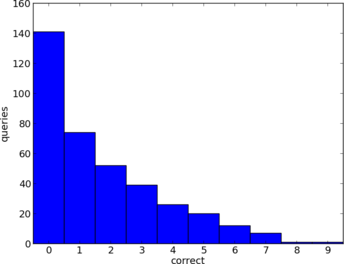}\\
    (a)&(b)&(c)
  \end{tabular}
  \caption{(a)~Average precision of the top 10 hits for the 141
    sentences as a function of the threshold on the sentence-tracker
    score.
    Without a threshold, (b)~the number of sentences with at most the given
    number of hits and
    (c)~the number of sentences with at least the given number of correct hits.}
  \label{fig:results}
\end{figure}

We generate 204 sentences that conform to the grammar in
Fig.~\ref{fig:grammar} from the following template:
\begin{small}
\begin{verbatim}
X {approached Y {,quickly,slowly} {,from the left,from the right},
   {lead,rode} Y {,quickly,slowly}
     {,leftward,rightward,{towards,away from} Z}}
\end{verbatim}
\end{small}
where X, Y, and Z are either \emph{person} or \emph{horse}.
We eliminate the 63 queries that involve people riding people and horses
riding people or other horses, as our video corpus has no positive examples for
these sentences.
This leaves us with 141 queries which conform to our grammar.
For each sentence, we score every video-sentence pair and return the
top 10 best-scoring clips for that sentence.
Each of these top 10 clips was annotated by a human judge with a
binary decision: is this sentence true of this clip?
In Fig.~\ref{fig:results}(a), we show the precision of the system on
the top 10 queries as a function of a threshold on the scores.
As the threshold nears zero, the system may return fewer than
10 results per sentence because it eliminates query results which are
unlikely to be true positives.
As the threshold tends to $-\infty$, the average precision across all
top 10 clips for all sentences is 20.6\%, and at its peak, the average
precision is 40.0\%.
In Fig.~\ref{fig:results}(b), we show the number of results returned
per sentence, eliminating those results which have a score of $-\infty$
since that tells us no tracks could be found which agree with the
semantics of the sentence,
On average, there are 7.96 hits per sentence, with standard deviation
3.61, and with only 14 sentences having no hits.
In Fig.~\ref{fig:results}(c), we show the number of correct his per
sentence.
On average, there are 1.63 correct hits per sentence, with standard
deviation 2.16, and with 74 sentences having at least one true positive.

We highlight the usefulness of this approach in
Fig.~\ref{fig:results-clips} where we show the top 6 hits for two
similar queries: \emph{The person approaches the horse} and \emph{The
  horse approached the person}.
Hits are presented in order of score, with the highest scoring hit in
the top left-hand corner and scores decreasing as one moves to the
right and to the next line.
Note how the results for the two sentences are very different from each other
and each sentence has 3 true positives and 3 false
positives.\footnote{\label{fn:video}It can be difficult to distinguish true
  positives and false positives from just a pair of frames, so we have included
  the full videos in the supplementary material along with additional results.}
With existing systems, both queries would provide the same hits as they
treat the sentences as conjunctions of words.

We compare against earlier work for two reasons even though we have established
that this approach is fundamentally more expressive.
First, to show that this approach is not only more expressive but buys you a
very large performance increase.
Second, to stave off a potential and natural objection to these results,
namely: surely because these are westerns if all you do is pick some movies at
random a person will be riding a horse in them.
This is not the case, movies are designed for their artistic qualities and
story-telling potential not their suitability as event recognition corpora and
the vast majority of scenes involve people talking to each other.
Rather than try to convince the reader of this we simply state that if this
task were easy, if there were so many true positives, surely the
state-of-the-art event recognition systems could find them?
This turns out to not be the case.

It's difficult to compare against earlier work because we must be fair to
that work.
Approaches like that of \cite{sivic2003video} use object detectors to find
nouns, but they do not use the state-of-the-art object detectors, like the
\cite{Felzenszwalb2010b} detector that we employ.
If we compared against this work we would be comparing two object detectors not
two retrieval approaches.
Moreover, systems which find events do not use the state-of-the-art event
recognition systems, hence this approach would dominate for unrelated reasons.
So we construct a system which follows the same principle as that of
\citet{christel2004exploiting}, \citet{worring2007mediamill}, and
\citet{snoek2007learned} but we build it out of the best components we can
find.
This system takes as input a query, two objects and a verb, note that this is
not a sentence as none of these approaches can truly encode the semantics of a
sentence.
We rank the videos by the average object detector score corresponding to the
participants.
We employ a recent event recognition system \cite{Kuehne11} to filter out this
ranking and keep videos which depict the query event.
This gives us a ranked list of videos, the better the object detector score the
more confident we are that the videos really do contain the query objects and
the event recognition system ensures the video depicts the query verb.
We train the event recognition system on the same 3 verbs, with 6 positive
examples per verb.
Note that our entire lexicon of 15 words was trained on 6 examples, 0.4
training samples per word, whereas we train the competing system on 15 times as
much training data.
We ran 3 queries, one for each verb, and each query was paired up with both
\emph{person} and \emph{horse}.
The results are as follows:
\begin{center}
  \begin{small}
    \begin{tabular}{@{}l|l|c|c@{}}
      our query & previous work query &  our TP & previous TP \\
      \hline
      The person rode the horse & \emph{person} \emph{horse} \emph{ride} & 9 & 0\\
      The person lead the horse & \emph{person} \emph{horse} \emph{lead} & 1 & 0\\
      The person approached the horse & \emph{person} \emph{horse} \emph{approach} & 4 & 1\\
    \end{tabular}
  \end{small}
\end{center}
where we measure the number of true positives in the top 10 hits for each
approach, and where a sample was deemed a true positive if it was described by
the query of the corresponding system.
Note that we far outperform this system which is an enhanced version of
existing work.

\begin{figure}
  \begin{tabular}{@{}c@{\hspace{1ex}}c@{}}
    \includegraphics[height=20ex]{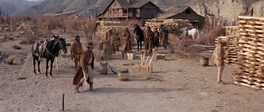}&
    \includegraphics[height=20ex]{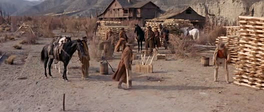}\\
    \includegraphics[height=20ex]{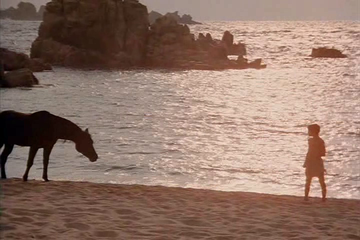}&
    \includegraphics[height=20ex]{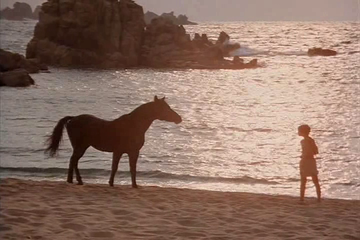}\\
    \includegraphics[height=20ex]{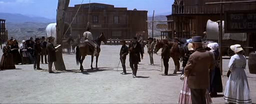}&
    \includegraphics[height=20ex]{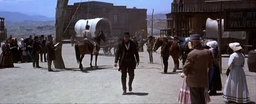}\\
    \includegraphics[height=20ex]{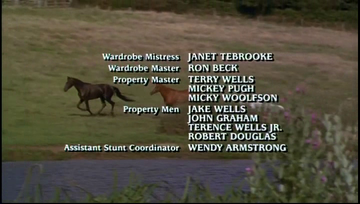}&
    \includegraphics[height=20ex]{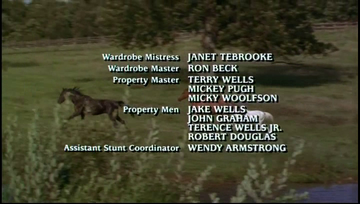}\\
    \includegraphics[height=20ex]{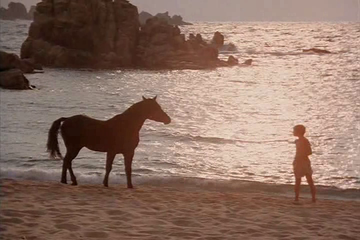}&
    \includegraphics[height=20ex]{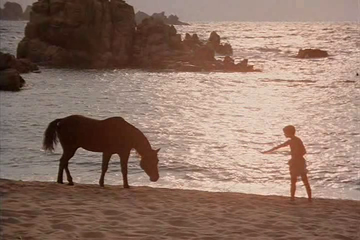}\\
    \includegraphics[height=20ex]{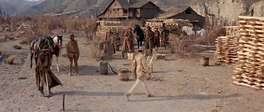}&
    \includegraphics[height=20ex]{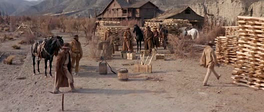}\\
  \end{tabular}
  \caption{The top 6 hits for the sentence \emph{The horse approached the person}.
    Half of the hits are true positives.}
  \label{fig:results-clips}
\end{figure}

\begin{figure}
  \begin{tabular}{@{}c@{\hspace{1ex}}c@{}}
    \includegraphics[height=20ex]{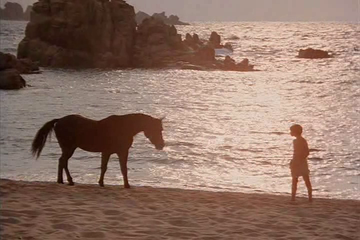}&
    \includegraphics[height=20ex]{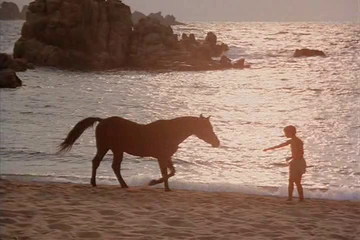}\\
    \includegraphics[height=20ex]{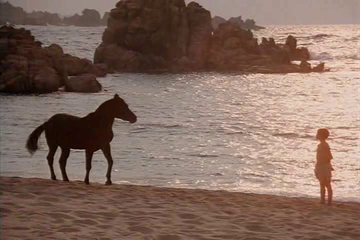}&
    \includegraphics[height=20ex]{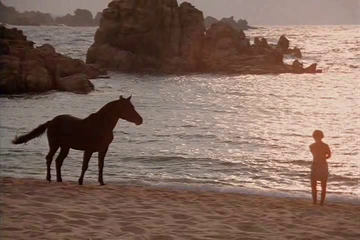}\\
    \includegraphics[height=20ex]{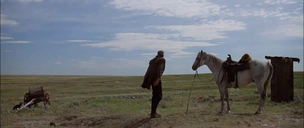}&
    \includegraphics[height=20ex]{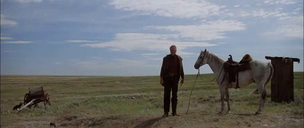}\\
    \includegraphics[height=20ex]{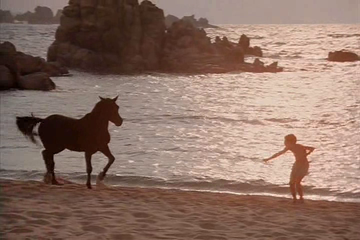}&
    \includegraphics[height=20ex]{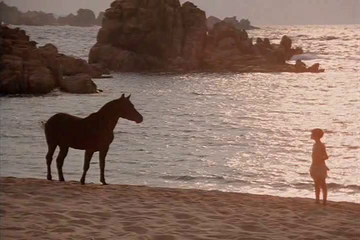}\\
    \includegraphics[height=20ex]{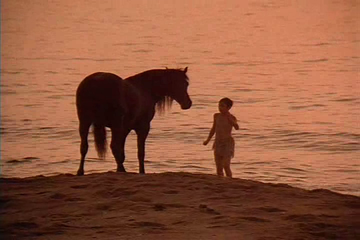}&
    \includegraphics[height=20ex]{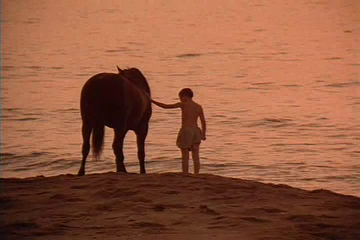}\\
    \includegraphics[height=20ex]{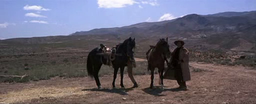}&
    \includegraphics[height=20ex]{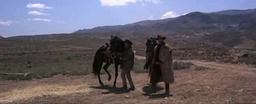}\\
    \multicolumn{2}{c}{(b) \emph{The person approached the horse}}
  \end{tabular}
  \caption{The top 6 hits for the sentence \emph{The person approached the horse}.
    Half of the hits are true positives.}
  \label{fig:results-clips}
\end{figure}

\section{Conclusion}

We have developed a framework for a novel kind of video search that
takes, as input, natural-language queries in the form of sentences, along
with a video corpus, and generates a list of ranked results.
This approach provides two novel video-search capabilities.
First, it can encode the semantics of sentences compositionally,
allowing it to express subtle distinctions such as the difference
between \emph{The person rode the horse} and \emph{The horse rode the
  person}.
Second, it can also encode structures more complex than just nouns and
verbs, such as modifiers, \eg\ adverbs, and entire phrases, \eg\
prepositional phrases.
We do not require any prior video annotation.
The entire lexicon shares a small number of parameters and, unlike
previous work, this approach does not need to be trained on every word
or even every related word.
We have evaluated this approach on a large video corpus of 10
Hollywood movies, comprising roughly 20 hours of video, by running 141
sentential queries and annotating the top 10 results for each query.

\ifnipsfinal
\subsubsection*{Acknowledgments}
This research was sponsored by the Army Research Laboratory and was
accomplished under Cooperative Agreement Number W911NF-10-2-0060.
The views and conclusions contained in this document are those of the authors
and should not be interpreted as representing the official policies, either
express or implied, of the Army Research Laboratory or the U.S. Government.
The U.S. Government is authorized to reproduce and distribute reprints for
Government purposes, notwithstanding any copyright notation herein.
\fi

\nocite{xu2005hmm,duan2005unified}

\bibliographystyle{abbrvnat}
\bibliography{arxiv2013b}

\begin{thebibliography}{20}
\providecommand{\natexlab}[1]{#1}
\providecommand{\url}[1]{\texttt{#1}}
\expandafter\ifx\csname urlstyle\endcsname\relax
  \providecommand{\doi}[1]{doi: #1}\else
  \providecommand{\doi}{doi: \begingroup \urlstyle{rm}\Url}\fi

\bibitem[Anjulan and Canagarajah(2009)]{anjulan2009unified}
A.~Anjulan and N.~Canagarajah.
\newblock A unified framework for object retrieval and mining.
\newblock \emph{IEEE Transactions on Circuits and Systems for Video
  Technology}, 19\penalty0 (1):\penalty0 63--76, 2009.

\bibitem[Aytar et~al.(2008)Aytar, Shah, and Luo]{aytar2008utilizing}
Y.~Aytar, M.~Shah, and J.~Luo.
\newblock Utilizing semantic word similarity measures for video retrieval.
\newblock In \emph{Proceedings of the IEEE Computer Society Conference on
  Computer Vision and Pattern Recognition}, pages 1--8, 2008.

\bibitem[Barbu et~al.(2012{\natexlab{a}})Barbu, Bridge, Burchill, Coroian,
  Dickinson, Fidler, Michaux, Mussman, Narayanaswamy, Salvi, Schmidt,
  Shangguan, Siskind, Waggoner, Wang, Wei, Yin, and Zhang]{Barbu2012a}
A.~Barbu, A.~Bridge, Z.~Burchill, D.~Coroian, S.~J. Dickinson, S.~Fidler,
  A.~Michaux, S.~Mussman, S.~Narayanaswamy, D.~Salvi, L.~Schmidt, J.~Shangguan,
  J.~M. Siskind, J.~W. Waggoner, S.~Wang, J.~Wei, Y.~Yin, and Z.~Zhang.
\newblock Video in sentences out.
\newblock In \emph{Proceedings of the Twenty-Eighth Conference on Uncertainty
  in Artificial Intelligence}, pages 102--112, 2012{\natexlab{a}}.

\bibitem[Barbu et~al.(2012{\natexlab{b}})Barbu, Siddharth, Michaux, and
  Siskind]{Barbu2012b}
A.~Barbu, N.~Siddharth, A.~Michaux, and J.~M. Siskind.
\newblock Simultaneous object detection, tracking, and event recognition.
\newblock \emph{Advances in Cognitive Systems}, 2:\penalty0 203--220,
  2012{\natexlab{b}}.

\bibitem[Chang et~al.(2002)Chang, Han, and Gong]{chang2002extract}
P.~Chang, M.~Han, and Y.~Gong.
\newblock Extract highlights from baseball game video with hidden {M}arkov
  models.
\newblock In \emph{Proceedings of the International Conference on Image
  Processing}, volume~1, pages 609--612, 2002.

\bibitem[Christel et~al.(2004)Christel, Huang, Moraveji, and
  Papernick]{christel2004exploiting}
M.~G. Christel, C.~Huang, N.~Moraveji, and N.~Papernick.
\newblock Exploiting multiple modalities for interactive video retrieval.
\newblock In \emph{Proceedings of the International Conference on Accoustics,
  Speech, and Signal Processing}, volume~3, pages 1032--1035, 2004.

\bibitem[Cooper et~al.(2007)Cooper, Liu, and Rieffel]{cooper2007video}
M.~Cooper, T.~Liu, and E.~Rieffel.
\newblock Video segmentation via temporal pattern classification.
\newblock \emph{IEEE Transactions on Multimedia}, 9\penalty0 (3):\penalty0
  610--618, 2007.

\bibitem[Duan et~al.(2005)Duan, Xu, Tian, Xu, and Jin]{duan2005unified}
L.-Y. Duan, M.~Xu, Q.~Tian, C.-S. Xu, and J.~S. Jin.
\newblock A unified framework for semantic shot classification in sports video.
\newblock \emph{IEEE Transactions on Multimedia}, 7\penalty0 (6):\penalty0
  1066--1083, 2005.

\bibitem[Everingham et~al.(2010)Everingham, Van~Gool, Williams, Winn, and
  Zisserman]{Everingham10}
M.~Everingham, L.~Van~Gool, C.~K.~I. Williams, J.~Winn, and A.~Zisserman.
\newblock The {PASCAL Visual Object Classes (VOC)} challenge.
\newblock \emph{International Journal of Computer Vision}, 88\penalty0
  (2):\penalty0 303--338, 2010.

\bibitem[Felzenszwalb et~al.(2010{\natexlab{a}})Felzenszwalb, Girshick, and
  McAllester]{Felzenszwalb2010b}
P.~F. Felzenszwalb, R.~B. Girshick, and D.~McAllester.
\newblock Cascade object detection with deformable part models.
\newblock In \emph{Proceedings of the IEEE Computer Society Conference on
  Computer Vision and Pattern Recognition}, pages 2241--2248,
  2010{\natexlab{a}}.

\bibitem[Felzenszwalb et~al.(2010{\natexlab{b}})Felzenszwalb, Girshick,
  McAllester, and Ramanan]{Felzenszwalb2010a}
P.~F. Felzenszwalb, R.~B. Girshick, D.~McAllester, and D.~Ramanan.
\newblock Object detection with discriminatively trained part based models.
\newblock \emph{{IEEE} Transactions on Pattern Analysis and Machine
  Intelligence}, 32\penalty0 (9):\penalty0 1627--1645, Sept.
  2010{\natexlab{b}}.

\bibitem[Hu et~al.(2011)Hu, Xie, Li, Zeng, and Maybank]{hu2011survey}
W.~Hu, N.~Xie, L.~Li, X.~Zeng, and S.~Maybank.
\newblock A survey on visual content-based video indexing and retrieval.
\newblock \emph{IEEE Transactions on Systems, Man, and Cybernetics, Part C:
  Applications and Reviews}, 41\penalty0 (6):\penalty0 797--819, 2011.

\bibitem[Kuehne et~al.(2011)Kuehne, Jhuang, Garrote, Poggio, and
  Serre]{Kuehne11}
H.~Kuehne, H.~Jhuang, E.~Garrote, T.~Poggio, and T.~Serre.
\newblock {HMDB}: a large video database for human motion recognition.
\newblock In \emph{Proceedings of the International Conference on Computer
  Vision}, 2011.

\bibitem[Sivic and Zisserman(2003)]{sivic2003video}
J.~Sivic and A.~Zisserman.
\newblock Video {G}oogle: A text retrieval approach to object matching in
  videos.
\newblock In \emph{Proceedings of the Ninth IEEE International Conference on
  Computer Vision}, pages 1470--1477, 2003.

\bibitem[Snoek et~al.(2007)Snoek, Worring, Koelma, and
  Smeulders]{snoek2007learned}
C.~G. Snoek, M.~Worring, D.~C. Koelma, and A.~W. Smeulders.
\newblock A learned lexicon-driven paradigm for interactive video retrieval.
\newblock \emph{IEEE Transactions on Multimedia}, 9\penalty0 (2):\penalty0
  280--292, 2007.

\bibitem[Viterbi(1971)]{Viterbi1971}
A.~J. Viterbi.
\newblock Convolutional codes and their performance in communication systems.
\newblock \emph{{IEEE} Transactions on Communication}, 19:\penalty0 751--772,
  Oct. 1971.

\bibitem[Worring et~al.(2007)Worring, Snoek, De~Rooij, Nguyen, and
  Smeulders]{worring2007mediamill}
M.~Worring, C.~G. Snoek, O.~De~Rooij, G.~Nguyen, and A.~Smeulders.
\newblock The mediamill semantic video search engine.
\newblock In \emph{Proceedings of the International Conference on Accoustics,
  Speech, and Signal Processing}, volume~4, pages 1213--1216, 2007.

\bibitem[Xu et~al.(2005)Xu, Ma, Zhang, and Yang]{xu2005hmm}
G.~Xu, Y.-F. Ma, H.-J. Zhang, and S.-Q. Yang.
\newblock An {HMM}-based framework for video semantic analysis.
\newblock \emph{IEEE Transactions on Circuits and Systems for Video
  Technology}, 15\penalty0 (11):\penalty0 1422--1433, 2005.

\bibitem[Yu and Siskind(2013)]{yu2013grounded}
H.~Yu and J.~M. Siskind.
\newblock Grounded language learning from video described with sentences.
\newblock In \emph{Proceedings of the 51st Annual Meeting of the Association
  for Computational Linguistics}, 2013.

\bibitem[Yu et~al.(2003)Yu, Xu, Leong, Tian, Tang, and Wan]{yu2003trajectory}
X.~Yu, C.~Xu, H.~W. Leong, Q.~Tian, Q.~Tang, and K.~W. Wan.
\newblock Trajectory-based ball detection and tracking with applications to
  semantic analysis of broadcast soccer video.
\newblock In \emph{Proceedings of the Eleventh ACM International Conference on
  Multimedia}, pages 11--20, 2003.

\end{thebibliography}

\end{document}